\documentclass{article}
\usepackage{ijcai21}

\usepackage{times}

\usepackage{soul}
\usepackage{url}
\usepackage[hidelinks]{hyperref}
\usepackage[utf8]{inputenc}
\usepackage[small]{caption}
\usepackage{graphicx}
\usepackage{amsmath}
\usepackage{booktabs}
\usepackage[utf8]{inputenc} 
\usepackage[T1]{fontenc}    
\usepackage{url}            
\usepackage{booktabs}       
\usepackage{amsfonts}       
\usepackage{nicefrac}       
\usepackage{microtype}      

\usepackage{amsmath}
\usepackage{amsthm}
\usepackage{amsfonts}

\newcommand{\app}{\raise.17ex\hbox{$\scriptstyle\sim$}}

\usepackage{graphicx}
\usepackage{blindtext}

\usepackage{floatrow}

\usepackage{graphicx}
\usepackage{wrapfig}

\usepackage[numbers]{natbib}

\theoremstyle{theorem}
\newtheorem{thrm}{Theorem}

\theoremstyle{theorem}

\usepackage{hyperref}       

\newcommand{\mb}[1]{\mathbf{#1}}
\newcommand{\bs}[1]{\boldsymbol{#1}}

\usepackage{tikz}
\usetikzlibrary{bayesnet}

\usepackage{amsthm}
\usepackage{amsmath}
\usepackage{amsfonts}
\usepackage{amssymb}
\usepackage{multirow}

\theoremstyle{definition}

\title{Few Shot Activity Recognition Using Variational Inference}

\author{
Neeraj Kumar$^1$\footnote{Contact Author}\and
Siddhansh Narang$^2$\\
\affiliations
$^1$Indian Institute of Technology, Delhi, India\\
$^2$RAeS (Royal Aeronautical Society), UK\\
\emails
neerajkr2k14@gmail.com,
siddhansh.narang@gmail.com
}

\begin{document}

\maketitle

\begin{abstract}
There has been a remarkable progress in learning a model which
could recognize novel classes with only a few labeled examples in the last few years. Few-shot learning (FSL) for action recognition is a challenging task of recognizing novel action categories which
are represented by few instances in the training data.

We propose a novel variational inference based architectural framework (HF-AR) for few shot activity recognition. Our framework leverages volume-preserving Householder Flow to learn a flexible posterior distribution of the novel classes. This results in better performance as compared to state-of-the-art few shot approaches for human activity recognition. approach consists of base model and an adapter model. Our architecture consists of a base model and an adapter model. The base model is trained on seen classes and it computes an embedding that represent the spatial an temporal insights extracted from the input video, e.g. combination of Resnet-152 and LSTM based encoder-decoder model. The adapter model applies a series of Householder transformations to compute a flexible posterior distribution that lends higher accuracy in the few shot approach. 

 Extensive experiments on three well-known datasets: UCF101, HMDB51 and Something-Something-V2, demonstrate similar or better performance on 1-shot and 5-shot classification as compared to state-of-the-art few shot approaches that use only RGB frame sequence as input. To the best of our knowledge, we are the first to explore variational inference along with householder transformations to capture the full rank covariance matrix of posterior distribution, for few shot learning in activity recognition.
\end{abstract}

\section{Introduction}

Action recognition in videos has received increasing attention from the deep learning community\cite{Wang2011ActionRB,Simonyan2014TwoStreamCN,Tran2015LearningSF,Carreira2017QuoVA} in recent years, due to its great potential value in industry and real-world applications.
Convolutional Neural Network (CNN) based methods \citep{Carreira2017QuoVA,Tran2015LearningSF} have achieved tremendous success in recognizing actions from videos in the supervised learning paradigm.

However, the performance of these methods deteriorates drastically \cite{xu2018dense} when recognizing action classes that are not adequately represented in the training data. This problem limits the deployment of these methods in real world applications where the number of action classes to be recognized increases rapidly with use cases.
In certain cases, the number of samples for novel classes are quite few for even traditional data augmentation techniques \cite{Krizhevsky2012ImageNetCW} to work. Therefore, systems with more advanced learning paradigm like Few Shot Learning \cite{Hariharan2016LowshotVO}, which learns to recognize novel classes from only a few examples (few-shots), have come into prominence.

Videos are more complex than images as one requires the complete information of the temporal aspects in recognizing some specific actions, such as opening the door.
 In the previous literature of video classification, 3D convolution and optical flow are two of the most popular methods to model temporal relations. The direct output of neural network encoders is always a temporal sequence of deep encoded features. Loss of information is even more severe for few-shot  learning. It is hard to learn the local temporal patterns which are useful for few-shot classification with limited amount of data. Utilizing the long-term temporal ordering information, which is often neglected in previous works on video classification, might potentially help with few-shot learning. For example, if the model could verify that there is
an act of playing long jump before a close-up view of a jumping person, the model will then become quite confident about predicting the class of this query video to be playing long jump, rather than some other potential predictions like jumping exercises or warm-up activities. In addition,
for two videos in the same class, even though they both contain the act of jumping, the exact temporal duration of each atomic step can vary dramatically. These \textit{non-linear temporal variations} in videos pose a significant challenge to few-shot video activity classification.

Prior research on few-shot learning for action recognition~\citep{mishra2018generative,guo2018neural,Zhu2018CompoundMN,dwivedi2019protogan,Zhang2020FewShotAR,Cao2020FewShotVC} leverage a variety of approaches, but either suffer from high computational complexity or are not able to deliver sufficient accuracy on datasets such as Something-Something-V2. \cite{mishra2018generative}, projects the the action class to visual space using word2vec, while \cite{guo2018neural} learns a graph based representation to classify the actions and \cite{Zhu2018CompoundMN} uses multi-saliency based approach to learn the video representation. \cite{dwivedi2019protogan} leverages a class-prototype vector with conditional GAN based approach to recognize the activity, while \cite{Zhang2020FewShotAR} uses C3D based features along with spatial and temporal network to detect unseen classes. Recently, \cite{Cao2020FewShotVC} used a temporal alignment network to generate the score used for few shot prediction. In an attempt to improve accuracy, we explore a novel approach using variational inference (\textbf{HF-AR}: Householder Flow \cite{Tomczak2016ImprovingVA}  for Activity Recognition) with normalizing flows and study the performance in detail on multiple well-known datasets. We demonstrate close or superior performance using a two stage variational inference architecture with a base and adapter model. The first stage consists of base model having pretrained Resnet-152\cite{He2016DeepRL} and stacked LSTM\cite{Hochreiter1997LongSM} layers to capture the spatial and temporal information and generate a relevant video embedding. The second stage is an adapter model that uses a series of householder transformations.

Specifically, we make the following contributions:\ 
\begin{itemize}
    \item We propose an architectural framework (HF-AR) that uses a base and an adapter model. The adapter model uses variational inference with volume preserving Householder Flow\cite{Tomczak2016ImprovingVA}. 
    \item This approach learns a full rank covariance matrix of the posterior distribution of datasets having novel classes  with few samples. This flexible posterior distribution helps in delivering higher accuracy in few shot video classification. 
    \item Detailed experimental analysis on three datasets namely: UCF101, HMDB51 and Something-Something-V2 demonstrates superior results for 5-shot classification as compared to state-of-the-art few shot approaches. 
    \item Ablation study on series of householder transformation demonstrates that the length of the householder transformation, $T$, depends on $k$ in the k-shot setting . Higher the value of $k$, more is the length required, $T$ to learn the posterior distribution to prevent over-fitting. .
\end{itemize}

\section{Related Work}
\paragraph{Few Shot Learning}
Few shot learning in image classification is predominantly based on meta-learning ~\citep{ML1,Zhang2018MetaGANAA,Finn2017ModelAgnosticMF}. Under the umbrella of meta-learning, various approaches like metric learning ~\cite{Snell2017PrototypicalNF,Vinyals2016MatchingNF,learntocpmpare}, learning optimization ~\cite{Ravi2017OptimizationAA} and learning to initialize and fine tune~\cite{Finn2017ModelAgnosticMF} are proposed.

 Metric learning methods learn the similarity between images~\cite{Vinyals2016MatchingNF} to classify samples of novel classes at inference based on nearest neighbors with labelled samples. Convolutional Siamese Net~\cite{Koch2015SiameseNN} employs a unique structure to naturally rank similarity between inputs. Once a network has been tuned, it can then capitalize on powerful discriminative features to generalize the predictive power of the network not just to new data, but to entirely new classes from unknown distributions.
Learning optimization involves memory based networks like LSTM~\cite{Xu2017FewShotOR} or memory augmented networks~\cite{Mishra2018ASN} which generate updates for the classifier rather than using gradients. The meta-learner also learns a task-common weight initialization which captures shared knowledge across tasks. Learning to initialize and fine-tune aims to make minimal changes in the network when adapting to the new task of classifying novel classes with few examples. Graph Neural Networks~\citep{gnn,Kim_2019_CVPR,Gidaris_2019_CVPR} construct a weighted graph to represent all the data and measure the similarity between data. 

Other methods use data augmentation, which learns to augment labeled data in unseen classes for supervised training~\citep{Hariharan2017LowShotVR,Wang2018LowShotLF}.
In this paper, we have propose a novel two-stage architecture, HF-AR, with a base and adapter model, in which the base model learns the knowledge (via embedding) from seen classes. This is then used in the adapter model that uses variational inference with a series of householder transformations, in the fine tuning phase to classify the novel classes.

\paragraph{Few shot Action Recognition:}
In contrast to few shot learning in images, few shot approaches in video classification have gotten less attention. In~\cite{mishra2018generative} each class attribute is projected to the visual space. In the visual space each class is represented by a Gaussian distribution. Our approach uses pre-trained models and  variational inference based architecture to generate embedding to predict the novel classes. Neural graph matching network~\cite{guo2018neural} used graph generator to leverage the structure of 3D data through a graphical representation. ~\cite{xu2018dense} combines dilated temporal convolution and densely layer connection to perform video action recognition.
\\

Compound Memory Networks~\cite{Zhu2018CompoundMN} introduces a multi-saliency embedding algorithm to encode video into a fixed-size matrix representation. Then, they propose a compound memory network (CMN) to compress and store the representation and classify videos by matching and ranking. Our method uses  pretrained Resnet-152 and LSTM layers to capture the spatial and temporal information to be used at fine tuning stage using variational inference. Attribute-based feature generation for unseen classes from GAN by using Fisher vector representation
was explored in zero-shot learning in ~\cite{Zhang2018VisualDS}. It proposes multi-level semantic inference to boost video feature synthesis, which captures the discriminative information implied in the joint visual-semantic distribution via feature-level and label-level semantic inference along with matching-aware Mutual Information Correlation to overcome information degradation issue. This captures seen-to-unseen correlation in matched and mismatched visual-semantic pairs by mutual information, providing the zero-shot synthesis procedure with robust guidance signals. Our approach, HF-AR, uses only RGB frames as input and captures the posterior distribution of novel classes, with few samples, using variational inference. 

ProtoGAN \cite{dwivedi2019protogan} proposes a GAN model which conditions CGAN on a class-prototype vector which is learnt through  Class Prototype Transfer Network (CPTN) to synthesize additional video features for the action recognition classifier. Action Relation Network~\cite{Zhang2020FewShotAR} contains C3D-based encoder, relation network and spatial and temporal attention units which refine the aggregation step.  ~\cite{Cao2020FewShotVC} uses  Temporal Alignment
Module (TAM), a novel few-shot learning framework that can learn to classify a previous unseen video. It computes temporal alignment score for each potential query support pair by averaging per-frame distances along a temporal alignment path, which enforces the score used to make prediction to preserve temporal ordering. Our approach (HF-AR) uses pretrained Resnet-152~\cite{He2016DeepRL} and stacked LSTM~\cite{Hochreiter1997LongSM} to capture the spatial and temporal information during pre-training on seen classes. During the fine tuning stage  on novel classes, we use a series of householder transformations to capture the full rank covariance matrix based distribution of the dataset.

\section{Mathematical background}
\subsection{Variational Autoencoder}

Let us consider some dataset $X = \{x^{(i)}\}_{i=1}^N$ consisting of $N$ i.i.d. samples of some continuous or discrete variable $x$. We assume that the data are generated by some random process, involving an unobserved continuous random variable $z$. The process consists of two steps: (1) a value $z^{(i)}$ is generated from some prior distribution $p(z)$; (2) a value $x^{(i)}$ is generated from some conditional distribution $p(x|z)$. Given $N$ data points $\mb{X} = \{\mb{x}_1, \ldots, \mb{x}_N\}$ we typically aim at maximizing the marginal log-likelihood:
\begin{equation}\label{eq:loglikelihood}
\ln p(\mb{X}) = \sum_{i=1}^{N} \ln p(\mb{x}_{i}),
\end{equation}
with respect to parameters. This task could be troublesome because the integral of the marginal likelihood $p(x) = \int p(z) p(x|z) \,dz$ is intractable (so we cannot evaluate or differentiate the marginal likelihood). These intractabilities are quite common and appear in cases of moderately complicated likelihood functions $p(x|z)$, e.g. a neural network with a nonlinear hidden layer. To overcome this issue one can introduce an \textit{inference model} (an \textit{encoder}) $q(\mb{z}|\mb{x})$ and optimize the variational lower bound:
\begin{equation}\label{eq:elbo}
\ln p(\mb{x}) \geq \mathbb{E}_{q(\mb{z}|\mb{x})}[ \ln p(\mb{x}|\mb{z}) ] - \mathrm{KL} \big{(} q(\mb{z}|\mb{x}) || p(\mb{z}) \big{)},
\end{equation}
where $p(\mb{x}|\mb{z})$ is called a \textit{decoder} and $p(\mb{z}) = \mathcal{N}(\mb{z}|\mb{0}, \mb{I})$ is the \textit{prior}. There are various ways of optimizing this lower bound but for continuous $\mb{z}$ this could be done efficiently through a re-parameterization of $q(\mb{z}|\mb{x})$ \citep{Kingma2014AutoEncodingVB, Rezende2014StochasticBA}. Then the architecture is called a \textit{variational auto-encoder} (VAE).

Typically, a diagonal covariance matrix of the encoder is assumed, \textit{i.e.}, $q(\mb{z}|\mb{x}) = \mathcal{N}\big{(} \mb{z}|\boldsymbol\mu (\mb{x}), \mathrm{diag}(\boldsymbol\sigma ^{2}(\mb{x})) \big{)}$, where $\boldsymbol\mu (\mb{x})$ and $\boldsymbol\sigma ^{2}(\mb{x})$ are parameterized by the NN. However, this assumption can be insufficient and not flexible enough to match the true posterior.

\subsection{Householder Flow}
A \textit{normalizing flow}, formulated by \citep{Tabak2013AFO, fg,Rezende2015VariationalIW} is a framework for designing flexible posterior distribution by starting with an initial random variable with a simple distribution for generating $\mb{z}^{(0)}$ and then applying a series of invertible transformations $\mb{f}^{(t)}$, for $t=1,\ldots , T$. Due to this , the last random variable $\mb{z}^{(T)}$ has more richer distribution. The aim of is to optimize the following objective function once we choose the transformations $\mb{f}^{(t)}$  for which we can compute the jacobian determinant.

\begin{multline}\label{eq:nfobjective}
\ln p(\mb{x}) \geq \mathbb{E}_{q(\mb{z}^{(0)}|\mb{x})} \Big{[} \ln p(\mb{x}|\mb{z}^{(T)}) + \sum_{t=1}^{T} \ln \Big{|}\mathrm{det}\frac{\partial \mb{f}^{(t)} }{ \partial \mb{z}^{(t-1)} } \Big{|} \Big{]} \\
- \mathrm{KL} \big{(} q(\mb{z}^{(0)}|\mb{x}) || p(\mb{z}^{(T)}) \big{)}.
\end{multline}

We have used the volume preserving flows that apply series of Householder transformations that we refer to as the \textit{Householder flow}. The volume preserving flows provide the series of transformation such that the jacobian determinant equals one. This helps in reducing the computation complexity while successfully providing a flexible posterior distribution.

\paragraph{Motivation:}

Any full-covariance matrix $\bs{\Sigma}$ can be represented by the eigenvalue decomposition using eigenvectors and eigenvalues:
\begin{equation}\label{eq:eigenvalue_decomposition}
\bs{\Sigma} = \mb{U} \mb{D} \mb{U}^{\top} , 
\end{equation}
where $\mb{U}$ is an orthogonal matrix with eigenvectors in columns, $\mb{D}$ is a diagonal matrix with eigenvalues. The goal is to model the matrix $\mb{U}$ to obtain a full-covariance matrix. This process requires a linear transformation of a random variable using an orthogonal matrix $\mb{U}$. Since the absolute value of the Jacobian determinant of an orthogonal matrix is $1$, for $\mb{z}^{(1)} = \mb{U} \mb{z}^{(0)}$ one gets $\mb{z}^{(1)} \sim \mathcal{N}( \mb{U}\bs{\mu}, \mb{U}\  \mathrm{diag}(\bs{\sigma}^{2})\ \mb{U}^{\top} )$. If $\mathrm{diag}(\bs{\sigma}^{2})$ coincides with true $\mb{D}$, then it would be possible to resemble the true full-covariance function.

Generally, the task of modelling an orthogonal matrix is rather non-trivial. However, first we notice that any orthogonal matrix can be represented in the following form \citep{Bischof1996OnOB, Sun1995ABR}:
\begin{thrm}\label{theorem:kernel_representation} \emph{(The Basis-Kernel Representation of Orthogonal Matrices)}\\
For any $M \times M$ orthogonal matrix $\mb{U}$ there exist a full-rank $M \times K$ matrix $\mb{Y}$ (the \textit{basis}) and a nonsingular (triangular) $K \times K$ matrix $\mb{S}$ (the \textit{kernel}), $K \leq M$, such that:
\begin{equation}
\mb{U} = \mb{I} - \mb{Y} \mb{S} \mb{Y}^{\top} .
\end{equation}
\end{thrm}
The value $K$ is called the \textit{degree} of the orthogonal matrix. Further, it can be shown that any orthogonal matrix of degree $K$ can be expressed using the product of Householder transformations \citep{Bischof1996OnOB, Sun1995ABR}, namely:
\begin{thrm}\label{theorem:number_of_householders}
Any orthogonal matrix with the basis acting on the $K$-dimensional subspace can be expressed as a product of exactly $K$ Householder transformations:
\begin{equation}
\mb{U} = \mb{H}_{K} \mb{H}_{K-1} \cdots \mb{H}_{1},
\end{equation}
where $\mb{H}_{k} = \mb{I} - \mb{S}_{kk} \mb{Y}_{\cdot k} (\mb{Y}_{\cdot k})^{\top}$, for $k=1,\ldots,K$.
\end{thrm}

Theoretically, Theorem \ref{theorem:number_of_householders} shows that we can model any orthogonal matrix using $K$ Householder transformations.

\begin{figure}[!htbp]
\includegraphics[width=1.0\textwidth]{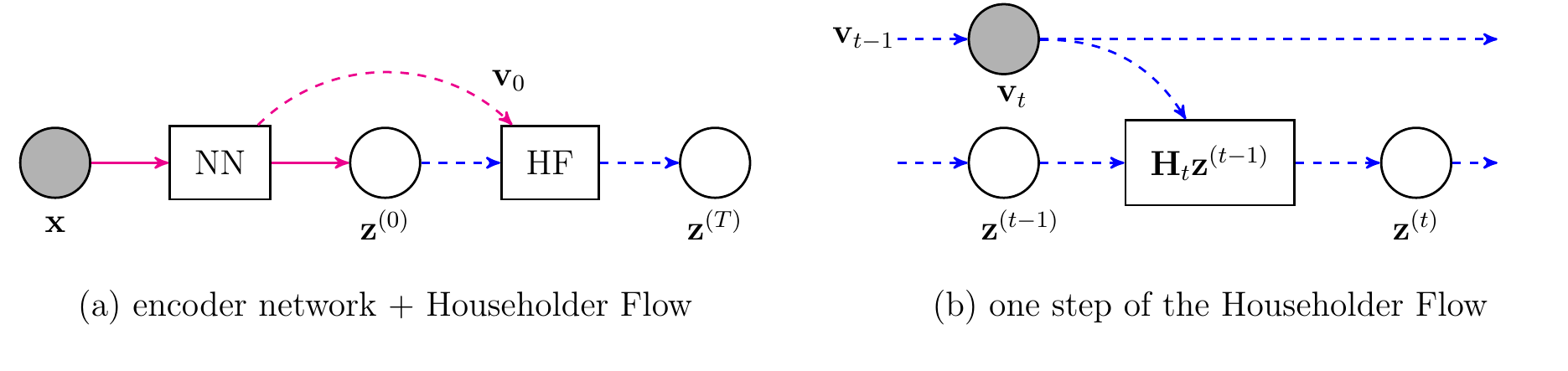}
\vskip -0.5cm
\caption{a. Figure shows the general architecture of VAE+HF b. Figure shows the series of housefolder transformation\cite{Tomczak2016ImprovingVA} }
\label{fig:householder_flow}
\end{figure}

\paragraph{Definition} 

For a given vector $\mb{z}^{(t-1)}$ the reflection hyperplane can be defined by a vector (a \textit{Householder vector}) $\mb{v}_{t} \in \mathbb{R}^{M}$ that is orthogonal to the hyperplane, and the reflection of this point about the hyperplane is \citep{Householder1958UnitaryTO}:
\begin{align}\label{eq:reflection}
\mb{z}^{(t)} &= \Big{(} \mb{I} - 2 \frac{\mb{v}_{t}\mb{v}_{t}^{\top}}{||\mb{v}_{t}||^{2}} \Big{)} \mb{z}^{(t-1)} \\
&= \mb{H}_{t} \mb{z}^{(t-1)},
\end{align}

where $\mb{H}_{t} = \mb{I} - 2 \frac{\mb{v}_{t}\mb{v}_{t}^{\top}}{||\mb{v}_{t}||^{2}}$ is called the \textit{Householder matrix}.

The series of such transformation results in $\ln \Big{|}\mathrm{det}\frac{\partial \mb{H}_{t}\mb{z}^{(t-1)} }{ \partial \mb{z}^{(t-1)} } \Big{|} = 0$, for $t=1,\ldots,T$. We start with the simple diagonal covariance matrix for $\mb{z}^{(0)}$ and after series of householder transformation results in vectors $\mb{v}_{t}$, $t=1,\ldots, T$, which leads to learn  full covariance matrix for more flexible posterior distribution. Figure \ref{fig:householder_flow} shows the idea of householder flow where where $T$ operations are required to generate flexible posterior distribution with approximated full covariance matrix.

\section{Architecture}

\begin{figure}[h!]
  \includegraphics[width=0.8\linewidth]{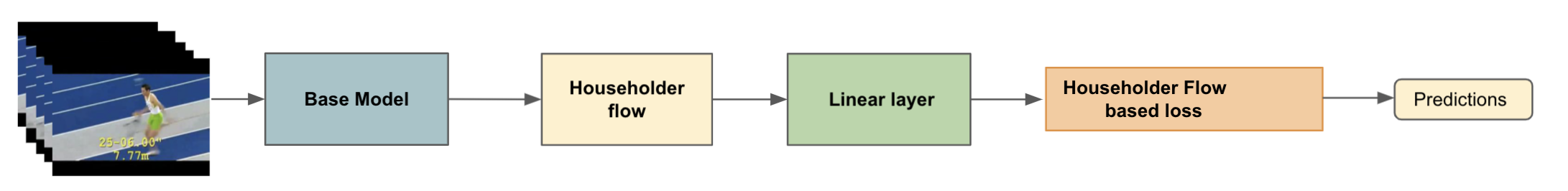}
  \caption{2-stage Proposed Architecture (HF-AR): Base and Adapter model. Base model with RGB frames as input and Adapter model with embeddings as input and applies a series of householder transformations.}
  \label{fig:adaptermodel}
\end{figure}

\begin{figure}[h!]
  \includegraphics[width=0.8\linewidth]{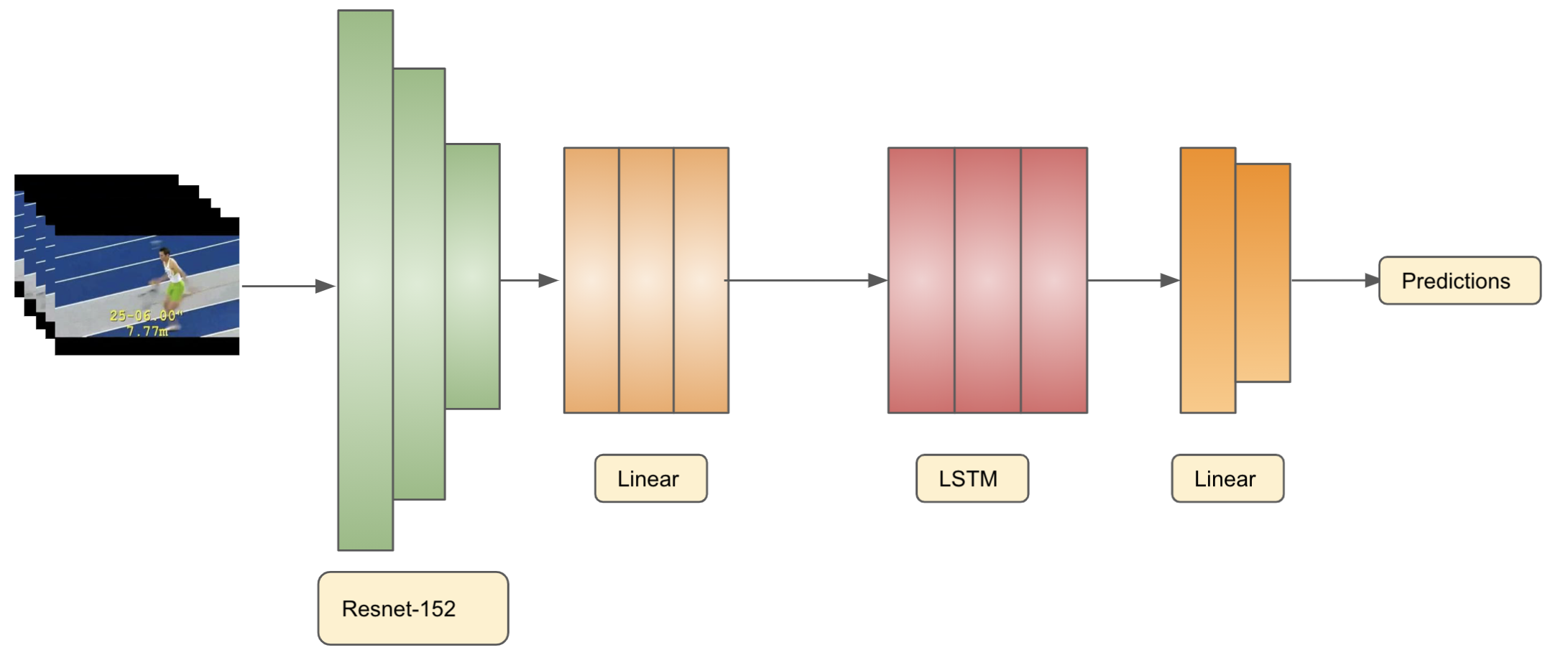}
  \caption{Base Model: takes video (RGB) frames as input. These are fed to pretrained Resenet-152 in the encoder followed by stacked LSTM layers in the decoder.}
  \label{fig:basemodel}
\end{figure}

Our proposed architecture (HF-AR) consists of two stages: base model and adapter model (Fig.~\ref{fig:adaptermodel}). The first stage (Fig.~\ref{fig:basemodel}) is an encoder-decoder based architecture to generate video embeddings on seen classes. The second stage is an adapter model which takes as input the embedding generated by the  base model and applies a series of householder transformations to finally predict the novel classes in a few shot fashion.

\subsection{Notation}
Let $x$, $\bar{x}$, $y$ represent real video frames,  video embeddings, class labels respectively. Let $\mathcal{S}$ = $\{x^s$, $y^s | x^s \in \mathcal{X}^s$, $y^s \in \mathcal{Y}^s\}$ be the training set for seen classes where $x^s$ denotes the  video frames , $y^s$ denotes the class labels in $\mathcal{Y}^s = \{y^{s}_1,\ldots,y^{s}_S\}$ with $S$ seen classes. The video embeddings $\bar{x}$ are calculated from the base model with $x^s$ video frames as an input. Additionally, $\mathcal{N} =\{x^n, y^n) | {x^n} \in \mathcal{X}^n, y^n \in \mathcal{Y}^n\}$ is available during adapter training from novel categories where $y^n$ is a class from a disjoint label set $\mathcal{Y}^n = \{y^{n}_1,\ldots,y^{n}_N\}$ of $N$ novel labels.  During the adaptor training , the video embeddings $\bar{x}$ are fed to the pretrained base model with multiple householder transformations to predict the novel classes in few shot setting. 

\subsection{Base Model}
The base model has encoder$-$decoder architecture in which the encoder takes video frames($x^s$) and feed it to the pretrained Resnet-152~\cite{He2016DeepRL} model followed by the $3$ linear layers to generate the $512$ dimensional video embedding($\bar{x^s}$). The decoder takes the embedding and send it to the stacked LSTM~\cite{Hochreiter1997LongSM} layer followed by linear layers to predict the seen classes. When the number of samples in the seen classes($y^s$) are high during the training, the encoder generates effective embeddings which helps the decoder to predict the seen classes with high accuracy. Pretrained Resnet-152 helps in exploiting the higher level features in the frames which compresses the information into small sized embeddings. The stacked LSTM helps in extracting the temporal aspects of the given input video which helps the base model to generate better embeddings. 

\subsection{Adapter Model}
The adapter model takes the video embedding as input and applies a series of volume preserving householder transformations to predict the novel classes. The householder transformation help the model to learn a more flexible posterior distribution of dataset with novel classes in few shot setting, with full rank covariance matrix . 

\subsection{Losses}
The optimisation function for variational inference is given in Equation~\ref{eq:nfobjective11}:
\begin{multline}
\label{eq:nfobjective11}
\ln p(\mb{x}) \geq \mathbb{E}_{q(\mb{z}^{(0)}|\mb{x})} \Big{[} \ln p(\mb{x}|\mb{z}^{(T)}) + \sum_{t=1}^{T} \ln \Big{|}\mathrm{det}\frac{\partial \mb{f}^{(t)} }{ \partial \mb{z}^{(t-1)} } \Big{|} \Big{]} \\
- \beta \cdot \mathrm{KL} \big{(} q(\mb{z}^{(0)}|\mb{x}) || p(\mb{z}^{(T)}) \big{)}.
\end{multline}

\paragraph{Cross Entropy loss}
In multi class classification, the class label($y^s$) follows the multinomial distribution given in Equation~\ref{eq:rhosum1e}. So the first part of  Equation \ref{eq:nfobjective11} i.e. $\ln p(\mb{x}|\mb{z}^{(T)})$  is the cross entropy loss in our set up. The cross entropy loss(CE) is given by Equation \ref{eq:ce} .

 \begin{equation}
 \label{eq:rhosum1e}
   \sum_{j=1}^{N}  y^{(j)} = 1
\end{equation}

  \begin{equation}
 \label{eq:ce}
   CE(\hat{y},y) = \sum_{j=1}^{N} y^{(j)} \log \hat{y}^{(j)}
\end{equation}

where $y^{(j)} $ is the true label i.e $0$ or $1$ and $\hat{y}^{(j)}$ is the predicted probability of a class coming from softmax operation.

\paragraph{Jacobian Determinant based loss} 
The second part, $\ln \Big{|}\mathrm{det}\frac{\partial \mb{H}_{t}\mb{z}^{(t-1)} }{ \partial \mb{z}^{(t-1)} } \Big{|}$ becomes zero as we are using the householder flow which is volume preserving based flow, whose jacobian determinant is 1.

\begin{table*}[t]
\caption{Comparison of accuracy between our model (HF-AR) and state-of-the-art approaches on HMDB51, UCF101 and  Something-Something V2 datasets}
\label{tab:compare}
\begin{tabular}{l|cccccc}
\toprule
\multirow{2}{*}{$\;\;\;\;\;\;$Model} & \multicolumn{2}{c}{HMDB51\cite{Kuehne2011HMDBAL}} & 
\multicolumn{2}{c}{UCF101\cite{Soomro2012UCF101AD}} &
\multicolumn{2}{c}{Something-SomethingV2 \cite{Goyal2017TheS}} \\
 & 1-shot & 5-shot & 1-shot & 5-shot & 1-shot & 5-shot \\ \hline
\textit{GenApp} $\;\;\;\;$~\cite{Mishra2018AGA} & ${- }$ & ${52.5 \pm 3.10}$ & ${-}$ & ${78.6 \pm 2.1}$ & $-$ & $-$ \\
\textit{ProtoGAN} ~\cite{dwivedi2019protogan} & $34.7 \pm 9.20 $ & $54.0 \pm 3.90$ & $57.8 \pm 3.0$ & $80.2 \pm 1.3$ & $-$ & $-$ \\
\textit{ARN}~\cite{Zhang2020FewShotAR} & $\mathbf{44.6 \pm 0.9}$ & $59.1 \pm 0.8$ & $\mathbf{62.1 \pm 1.0}$ & $84.8 \pm 0.8$ & $-$ & $-$ \\
\textit{TAM} $\;\;\;\;$~\cite{Cao2020FewShotVC} & ${- }$ & ${-}$ & ${-}$ & ${-}$ & $42.8$ & $52.3$ \\
\hline
\textit{HF-AR} & $43.4 \pm 0.6$ & $\mathbf{62.2 \pm 0.9}$ & $58.6\pm 1.2$ & $\mathbf{86.4 \pm 1.6}$ & $\mathbf{43.1 \pm 1.7}$ & $\mathbf{55.13 \pm 3.1}$ \\
\bottomrule
\end{tabular}
\end{table*}

\paragraph{Kullback–Leibler(KL) divergence loss}
The third part, $\mathrm{KL} \big{(} q(\mb{z}^{(0)}|\mb{x}) || p(\mb{z}^{(T)}) \big{)}$ is the KL loss between with $q(\mb{z}^{(0)}|\mb{x})$ and $p(\mb{z}^{(T)})$ where $\mb{z}^{(T)}$ is the last random variable obtained from householder flow. The KL loss acts as the regularizer which prevents the model from overfitting and helps in learning a flexible posterior distribution. $\beta$ value is used to to control the effect of KL loss. The value of $\beta$ decreases linearly with every epoch.

\section{Experimental Section}
In this work, our task is few-shot video classification, where the objective is to classify novel classes with only a few examples from the support set.

\subsection{Datasets}
We evaluate our method (HF-AR) on three publicly available datasets for action recognition. On these datasets, we measure the action recognition performance with respect to top-1 and top-5 recognition accuracy.
\begin{itemize}
    \item UCF101\cite{Soomro2012UCF101AD}: contains $13320$ videos spanning over $101$ classes. For our experiments, we randomly split the data into $80$ seen and $21$ novel classes.
    \item HMDB51\cite{Kuehne2011HMDBAL}: contains $6766$ videos spanning over $51$ classes. For our experiments we randomly split the data into $41$ seen and $10$ novel classes.
    \item Something-Something V2~\cite{Goyal2017TheS}: We randomly selected $100$ classes from the whole dataset. For our experiments, we randomly split this
data into $76$ seen and $24$ novel classes.
\end{itemize}

\subsection{Implementation Details}

\paragraph{Pre-Processing Steps}: The videos frames are resized into 224*224 and then normalized before being fed to the activity recognition model. We have selected $50$ frames of the video to be used in the proposed model.
The training procedure uses the batch of $150$ during base model training and $4$ during adapter training . We have used Adam optimizer~\cite{Kingma2015AdamAM} with the initial learning rate of $1e-3$ and dynamic adjustment. We have used ReduceLROnPlateau scheduler~\cite{Mukherjee2019ASD} with patience equals to $10$.

\subsection{Model Details}
 Base model consists of encoder decoder framework . The encoder uses the pretrained Resnet 152 which gives $2048$ features for every frame and finally all the $50$ frames are concatenated to generate the tensor of size $(batchsize,50,2048)$. Three linear layers are used  with channel length of $(2048 \rightarrow 300)$ , $(512 \rightarrow 512)$, $(512 \rightarrow 512)$. The $300$ embeddings of every frames are fed into the decoder architecture of the base model.

The decoder takes the tensor of size $(batchsize,50,512)$ and feeds this to $3$ stacked LSTM layers. The output of the last layer of LSTM is used as an input to the following layers. Two linear layers having channel length $(512 \rightarrow 256)$ ,$(256 \rightarrow classes)$ are then used to predict the classes. 

The adapter model which is used to predict the classes with few samples uses the pretrained base model as the source of its input. The pretrained model gives the output of $256$ features (output of second last layer of base model) is fed to the householder flow network which predicts the classes. The householder network uses linear layers to predict the $\mu$ and $\sigma$ and which generates the vector using the reparametrization trick, $\mb{z}^{(0)}$= $( \mu + \sigma\odot\epsilon)$. The $\mb{z}^{(0)}$ goes into series of householder transformations to generate the final $\mb{z}^{(T)}$ whose feature size is $256$. This goes into linear layer to predict the novel classes having few samples.

\subsection{Comparative Analysis}
    Table~\ref{tab:compare} presents the comparison of accuracy obtained between our approach (HF-AR) vs. state-of-the-art approaches on the three datasets with 1-shot and 5-shot classification. For 5-shot classification, HF-AR delivers better accuracy on all the 3 datasets: $62.2\%$ vs $59.1\%$ (ARN~\cite{Zhang2020FewShotAR}) for HMDB51 dataset, $86.4\%$ vs $84.8\%$ (ARN~\cite{Zhang2020FewShotAR}) for UCF101 dataset, and $55.13\%$ vs $52.3\%$ (TAM~\cite{}) for Something-Something V2 dataset. This improvement is due to flexbile posterior model learned by Householder Flow in our adapter model. For 1-shot classification, HF-AR delivers slightly better accuracy for Something-Something V2 dataset: $43.1\%$ vs $42.8\%$ (~\cite{Cao2020FewShotVC}); slightly lower accuracy for HMDB51 dataset: $43.4\%$ vs $44.6\%$ (ARN~\cite{Zhang2020FewShotAR}); and, lower accuracy for UCF101 dataset: $58.6\%$ vs $62.1\%$ (~\cite{Cao2020FewShotVC}). For very limited data as in 1-shot approach, we believe it becomes harder for our variational inference based approach to attain the right generalization ability.

\subsection{Ablation Study}

\paragraph{Effect of length of householder flow}: We examined the behaviour of the length of the series of householder transformations in the few shot setting. We have done experiments with HMDB51 dataset to study this in detail. Table~\ref{tab:incr} shows that in 1-shot setting as we increase the length, the accuracy dips as the model is overfitting. In 5-shot setting the optimum length increases as compared to 1-shot setting.

\begin{table}[h!]
    \begin{tabular}{c|c|c} 
      \textbf{Length} & \textbf{1-shot} & \textbf{5-shot} \\
      \hline
       T=1  & $35.1 \pm 0.6$ & $55.6 \pm 0.8$  \\
       T=3    & $\mathbf{43.4 \pm 0.6}$ & $59.9 \pm 0.8$  \\
       T=10  & $39.8 \pm 0.4$ &$\mathbf{ 62.2 \pm 0.9}$ \\
       T=20 & $39.2 \pm 0.4$ & $60.8 \pm 0.7$ \\
    \end{tabular}
    
    \vspace {0.25\baselineskip}
    \caption{Effect of changing the length of householder flow in HF-AR on accuracy with HMDB51 daatset.  }
    \label{tab:incr}
\end{table}

\begin{table}[h!]
    \begin{tabular}{c|c|c} 
      \textbf{Length} & \textbf{1-shot} & \textbf{5-shot} \\
      \hline
       3D-CNN \cite{Tran2015LearningSF} & $45.4 \pm 1.6$ & $70.5 \pm 1.3$  \\
       CompHF   & $48.4 \pm 2.3$ & $72.2 \pm 2.1$  \\
       HF-AR  & $\mathbf{58.6 \pm 1.2}$ & $\mathbf{86.4 \pm 1.6}$ \\
    \end{tabular}
    
    \vspace {0.25\baselineskip}
    \caption{Effect of various Base Model with UCF101 dataset on accuracy.  }
    \label{tab:base-abalation}
\end{table}

\paragraph{Effect of various base models}: We studied the effect of various base models in few shot setting with UCF101 dataset. Table~\ref{tab:base-abalation} shows superior performance of our proposed model with respect to other base models. 3D-CNN ~\cite{Tran2015LearningSF} based base model is not able to predict high quality video embeddings which leads to lower accuracy on both seen and unseen classes. However, our proposed method (HF-AR) uses pretrained ResNet-152 which is able to learn good image features along with LSTM which captures the temporal aspects. The CompHF base model uses householder flow in the base model itself to predict the seen classes. It gives lesser accuracy on seen and unseen classes because the posterior distribution generated from householder flow fails to capture the more complex multi-modal distribution of the dataset with larger number of seen classes. This needs further investigation with more variety of base models. 

\section{Conclusions}
We propose a variational inference based architectural framework for few shot activity recognition. The proposed architecture does not require any optical flow for input or supervision and provides a unified and efficient  method to predict unseen classes. This method learns the full rank covariance matrix of the posterior distribution of datasets with novel classes. The idea relies on the observation that the true full-covariance matrix can be decomposed to the diagonal matrix with eigenvalues on the diagonal and an orthogonal matrix of eigenvectors that could be further modeled using a series of Householder transformations. Experimental study on multiple datasets demonstrates that variational inference with Householder flows could perform better than other state of the art approaches for few shot activity recognition. This makes this approach a very promising direction for further investigation in this and other related areas in computer vision.


\end{document}